%% file: 00-main.tex
\documentclass[preprint,review,12pt]{elsarticle}

\usepackage{lineno,hyperref}
\modulolinenumbers[5]

\usepackage{comment}
\usepackage{enumitem}
\usepackage{graphicx}
\usepackage{subfig}
\usepackage{multicol}
\usepackage{multirow}
\usepackage[dvipsnames]{xcolor}
\usepackage{algorithm}
\usepackage{algorithmicx}
\usepackage[noend]{algpseudocode}
\usepackage{wrapfig,lipsum,booktabs}
\usepackage{rotating}
\usepackage{lscape}

\algnewcommand{\LeftComment}[1]{\State \(\triangleright\) #1}

\newcommand{\eg}{{\textit{e.g.}}}
\newcommand{\ie}{{\textit{i.e.}}}
\newcommand{\etal}{{\textit{et al.}}}

\usepackage{array}
\newcommand{\PreserveBackslash}[1]{\let\temp=\\#1\let\\=\temp}
\newcolumntype{C}[1]{>{\PreserveBackslash\centering}p{#1}}

\journal{}

\begin{document}

\begin{frontmatter}

\title{Online Evolutionary Neural Architecture Search for Multivariate Non-Stationary Time Series Forecasting}

\tnotetext[mytitlenote]{This material is based upon work supported by the U.S. Department of Energy, Office of Science, Office of Advanced Combustion Systems under Award Number \#FE00031750 and \#FE0031547. It is also supported by the Federal Aviation Administration and MITRE Corporation under the National General Aviation Flight Information Database (NGAFID) award. This work has been partially supported by the National Science Foundation under Grant Number 2225354. Any opinions, findings, and conclusions or recommendations expressed in this material are those of the authors and do not necessarily reflect the views of the National Science Foundation.}


\author{Zimeng Lyu\corref{mycorrespondingauthor}}
\cortext[mycorrespondingauthor]{Corresponding author}

\author{Alexander Ororbia}
\author{Travis Desell}


\address{Rochester Institute of Technology, Rochester, NY, USA}

\input{01-abstract}



\end{frontmatter}

\input{02-introduction}
\input{03-relatedwork}
\input{04-method}

\input{05-results}
\input{06-conclusion}

\bibliography{99-reference}

\end{document}

%% file: 01-abstract.tex
\begin{abstract}

Time series forecasting (TSF) is one of the most important tasks in data science given the fact that accurate time series (TS) predictive models play a major role across a wide variety of domains including finance, transportation, health care, and power systems. Real-world utilization of machine learning (ML) typically involves (pre-)training models on collected, historical data and then applying them to unseen data points. However, in real-world applications, time series data streams are usually non-stationary and trained ML models usually, over time, face the problem of data or concept drift.  

To address this issue, models must be periodically retrained or redesigned, which takes significant human and computational resources. Additionally, historical data may not even exist to re-train or re-design model with. As a result, it is highly desirable that models are designed and trained in an online fashion. This work presents the Online NeuroEvolution-based Neural Architecture Search (ONE-NAS) algorithm, which is a novel neural architecture search method capable of automatically designing and dynamically training recurrent neural networks (RNNs) for online forecasting tasks. Without any pre-training, ONE-NAS utilizes populations of RNNs that are continuously updated with new network structures and weights in response to new multivariate input data. ONE-NAS is tested on real-world, large-scale multivariate wind turbine data as well as the univariate Dow Jones Industrial Average (DJIA) dataset. Results demonstrate that ONE-NAS outperforms traditional statistical time series forecasting methods, including online linear regression, fixed long short-term memory (LSTM) and gated recurrent unit (GRU) models trained online, as well as state-of-the-art, online ARIMA strategies.

\end{abstract}

\begin{keyword}
Time Series Forecasting, 
Online Learning, 
Neural Architecture Search,
Recurrent Neural Networks,
Neuroevolution
\end{keyword}

%% file: 02-introduction.tex
\section{Introduction}
\label{sec:tinro}

Time series forecasting (TSF) is commonly used in many domains, such as health care~\cite{zinouri2018modelling}, transportation~\cite{wu2012online}, finance~\cite{cao2019financial}, and power systems~\cite{lyu2021neuroevolution}. TSF models are usually designed and trained offline with historical time series data. However, offline model building and training for TSF applications is based on the assumption that the target time series is stationary and that the models are to be trained with stationary data. When these pre-trained models are later applied to unseen temporal data, if the underlying data distribution of the data points change over time, these predictive systems begin to break down ~\cite{guo2016robust}~\cite{fields2019mitigating}. In real world TSF applications, time series data is usually non-stationary and suffers from data drift. Some applications rely on auxiliary methods to transform the data's non-stationary distribution to a stationary one by training the models in a batch manner (by periodically updating the model with new data) to maintain expected prediction accuracy. Unfortunately, this in turn can make models susceptible to catastrophic forgetting \cite{mccloskey1989catastrophic,french1999catastrophic} if the wrong historical data is removed in future training runs.

The increase in computational ability of personal computers and the accessibility of cloud computing makes it possible to do online TSF for various domains, such as the internet of things~\cite{raju2020iot}, climate modeling~\cite{partee2022using}, financial decision making~\cite{gonzalez2019fuzzy}, and power and energy systems~\cite{lyu2022one}. While traditional batch training methods can struggle to keep up with the large scale of modern  streaming data, online TSF methods offer the potential of real-time model updates.  Additionally, it may not be computationally feasible to examine model architecture changes when doing batch updates as it is not typically possible to do transfer learning between different model architectures.

However, online TSF faces a lot of challenges. The first problem is catastrophic forgetting -- when there is data drift in online TSF data, models tends to forget historical information~\cite{hoi2021online, yu2007online}. Even though rehearsal techniques can help to avoid catastrophic forgetting~\cite{french1999catastrophic}, they can increase model complexity and reduce prediction efficiency. To ensure online methods can handle real time data, tradeoffs need to be made between model complexity, learning capability, and efficiency. Additionally, real-world time series data tends to be multivariate, which adds more complexity due to a higher dimensionality, as well as non-seasonal.  This makes it challenging to effectively use classical statistical methods such as the autoregressive moving average (ARMA)~\cite{cryer1986time} and  autoregressive integrated moving average (ARIMA)~\cite{liu2016online} for online multivariate time series forecasting. Lastly, real-world, large-scale online data is usually noisy, and sometimes has incomplete data (\eg, when sensors may not be available), and TSF models need to be robust enough to deal with incomplete data streams. Models in this domain should also be able to adapt to multiple applications~\cite{hoi2021online}.


A major limitation for traditional and current online TSF models is that they assume a single model or neural network architecture which remains fixed, even as it is retrained on new batches of data or incoming online data. Due to this, a model for online TSF will either need to be large enough to be able to capture all possible aspects of concept drift, which is computationally inefficient/intractable, or  suffer from catastrophic forgetting when it is trained on new data, limiting its ability to predict data similar to historical observations. In order to overcome these limitations, based on our previous work with time series forecasting on real-world TSF data with neuroevolution (NE)~\cite{lyu2021experimental,lyu2021Improving,lyu2021neuroevolution}, we have developed the Online NeuroEvolution-based Neural Architecture Search (ONE-NAS) algorithm. ONE-NAS, to the authors' knowledge, is the first algorithm capable of training and evolving recurrent neural networks (RNNs) architectures in an online manner for TSF. Models architectures evolved through ONE-NAS can be quickly trained in response to incoming data streams, which stands in contrast to alternative methods that require significant offline  training time using previously gathered sets of training data. Furthermore, ONE-NAS dynamically adapts architectures to keep computational costs minimal while being more robust to catastrophic forgetting.

The novel contributions of our proposed ONE-NAS framework include:
\begin{itemize}[noitemsep,nolistsep]
  \item it is the first online NE neural architecture search (NAS) algorithm for TSF with RNNs, 
  \item it utilizes distributed islands that allow evolution in small niches,
  \item it utilizes island repopulation to improve performance over traditional methods,
  \item it generates genomes (RNN architectures and weights) using a Lamarckian weight initialization strategy that allows information retention (significantly speeding up training),
  \item it trains genomes on randomly selected historical data, such that important historical data is naturally preserved in the training process,
  \item it performs well on univariate as well as multivariate datasets,
  \item it is distributed and scalable, and, 
  \item it is robust to real-world, large scale, and noisy data.
\end{itemize}

\begin{figure*}
    \centering
	\includegraphics[width=\textwidth]{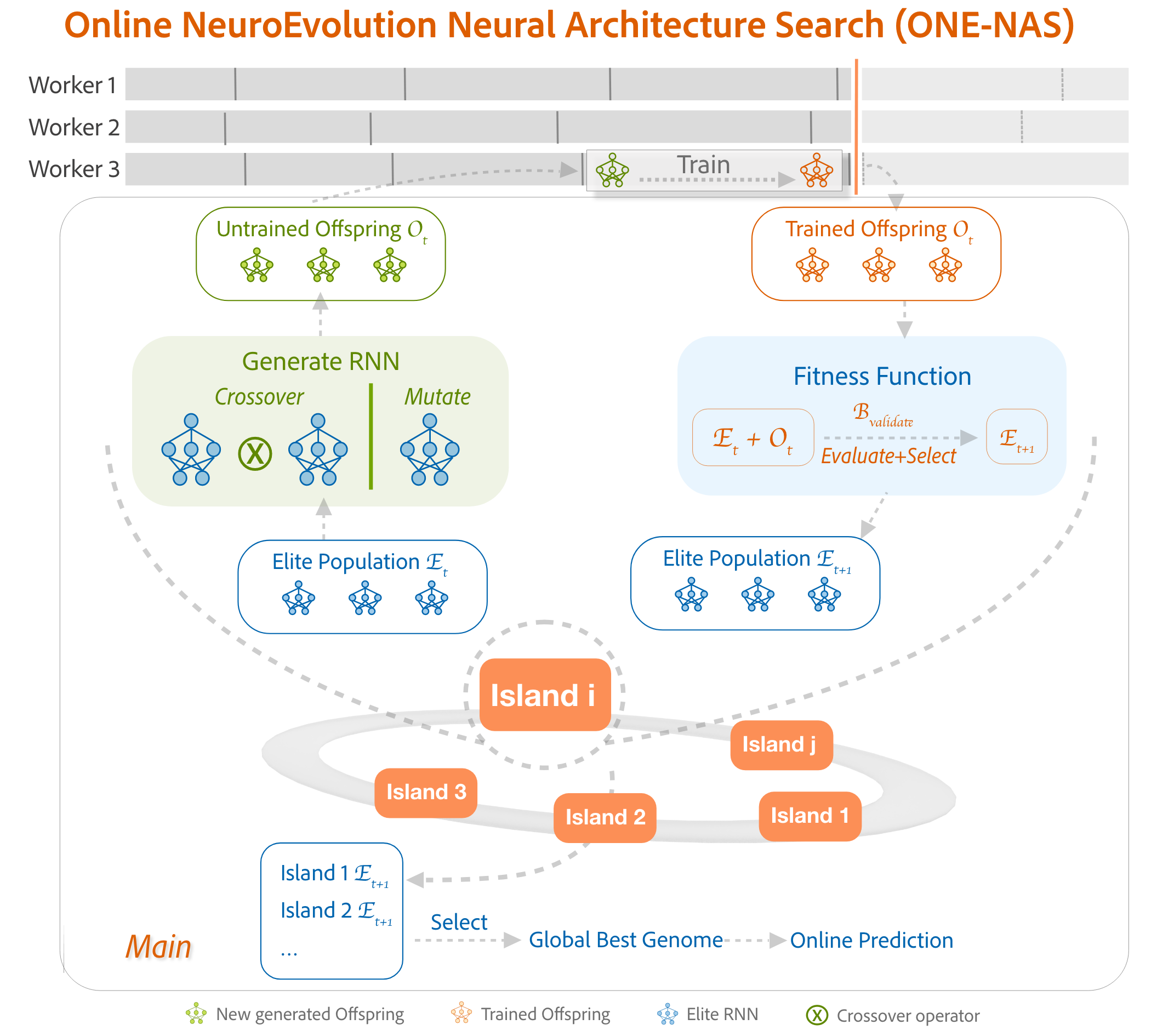}
	\caption{Depicted is one generation of the ONE-NAS online neuroevolutionary process. The global best genome performs predictions concurrently with distributed genome generation and evaluation. }
	\label{fig:onenas}
\end{figure*}

Preliminary work~\cite{lyu2022one}~only offered a minimal presentation/sketch of the ONE-NAS algorithm, and this work extends this by providing full treatment of the algorithm, including detailed pseudocode. Preliminary, earlier work further only compared the performance between ONE-NAS and classical naive, moving average, and exponential smoothing methods, focusing on the performance between ONE-NAS and ONE-NAS utilizing a repopulation method. This paper presents additional experiments comparing ONE-NAS with a traditional statistical TSF methods for online linear regression, online trained fixed architecture LSTM and GRU RNNs, as well as modern online ARMIA based methods -- crucially highlighting its feasibility as an algorithm that can both design and train RNNs for TSF in online scenarios.

Results were gathered using noisy, real-world multivariate time series data extracted from wind turbine sensors as well as the univariate Dow Jones Industrial Average (DJIA) dataset. Our empirical results demonstrate significant improvements in accuracy when using ONE-NAS over these methods. Furthermore, ONE-NAS utilizes a distributed, scalable computational processing scheme and is shown to operate efficiently, in real-time, over short time scales. Furthermore, we show that, within ONE-NAS, utilizing multiple islands that are periodically repopulated to prevent stagnation significantly improves the performance of the underlying optimization process.

%% file: 03-relatedwork.tex
\section{Related Work}
\label{sec:lit_review}
Classical time series forecasting methods include na\"ive prediction, moving average (MA) prediction~\cite{cryer1986time}, autoregression (AR)~\cite{stine1987estimating}, and exponential smoothing~\cite{gardner1985exponential} (also called simple exponential smoothing), offering powerful tools for time series forecasting; and traditional statistical methods such as autoregressive moving average (ARMA)~\cite{cryer1986time}, autoregressive integrated moving average (ARIMA) models~\cite{cryer1986time}, triple exponential smoothing~\cite{hansen1970triple} (also known as Holt Winter’s Exponential Smoothing) still play an important role in time series forecasting and have been used in a variety of different datasets~\cite{siregar2017comparison}~\cite{chen2019state}~\cite{benvenuto2020application}.  

Recent research has led to the development of online variants of classical models allowing model parameters to adapt to incoming streams of data. The online autoregressive integrated moving average (online ARIMA) has been proposed for online time series forecasting \cite{liu2016online}, anomaly detection \cite{kozitsin2021online}, and unsupervised anomaly detection~\cite{schmidt2018unsupervised}~\cite{schmidt2018unsupervised}. The autoregressive moving average (ARMA) \cite{anava2013online} and the seasonal autoregressive integrated moving average (SARIMA) \cite{han2018multivariate} have also been proposed as powerful models for time series, with Anava~\etal further proposing an AR model for TSF that can handle missing data values~\cite{anava2015online}. 

With respect to the artificial neural network (ANN) based approaches for online TSF, Guo~\etal~proposed an adaptive gradient learning method for training recurrent neural networks (RNNs) capable of time series forecasting notably in the presence of anomalies and change points~\cite{guo2016robust}. Yang~\etal~use RoAdam (Robust Adam) to train long short-term memory (LSTM) RNNs for online time series prediction in the presence of outliers~\cite{yang2017robust}. Wang~\etal~design an online sequential extreme learning machine utilizing a kernel (OS-ELMK) for non-stationary time series forecasting \cite{wang2014online}. Other than classical or NN-based online algorithms, online neural architecture search (NAS) and AutoML algorithms could also solve the drift problem. Some AutoML frameworks are designed to automatically adapt to data drift problem for online time series classification problems ~\cite{celik2022online}~\cite{celik2021adaptation}~\cite{madrid2019towards}. Yan~\etal~proposed a privacy-preserving online AutoML for face detection that extracts the meta-features of the input data with the goal of continuously improving the core algorithm's performance~\cite{yan2022privacy}. 

Neuroevolution (NE) itself has been widely used for time series prediction and neural architecture search in offline scenarios~\cite{ororbia2018using}~\cite{lyu2021Improving}~\cite{lyu2021experimental}. However, online neuroevolution has only been rarely investigated, with a few algorithms designed for games or simulators that involve real-time interactions, such as an online car racing simulator~\cite{cardamone2010learning}, online video games~\cite{agogino2000online}~\cite{stanley2005real}, and robotic controllers~\cite{galassi2016evolutionary}. Crucially, these online NE NAS algorithms are based on the venerable NeuroEvolution of Augmenting Topologies (NEAT) algorithm~\cite{stanley2002evolving} and start with minimal networks and evolve topologies and weights through a simulated evolutionary process.
Cardamone~\etal~developed an online car racing simulator based on NEAT~\cite{stanley2002evolving} and rtNEAT \cite{stanley2005real}, combined with four evaluation strategies ($\epsilon$-greedy~\cite{sutton2018reinforcement},
$\epsilon$-greedy-improved, softmax, and interval-based). This algorithm evolves car drivers from scratch and can outperform offline models. 
To our knowledge, none of the above online neuroevolution algorithms are capable of evolving recurrent networks nor have any been developed to specifically conduct online time series forecasting, notably making ONE-NAS the first of its kind.

%% file: 04-method.tex
\section{Methodology}
\label{sec:methodology}

This work leverages components from the Evolutionary Exploration of Augmenting Memory Models (EXAMM) algorithm~\cite{ororbia2019examm} at its core. In particular, our algorithm leverage's EXAMM's mutation, crossover, and training operations while maintaining the population in distinct islands. However, the process by which it evolves artificial neural networks (ANNs) and the operations it provides for utilizing them in online scenarios are novel. This section first describes EXAMM and its leveraged components in detail before describing how they are incorporated into and extended/generalized in ONE-NAS.

\subsection{EXAMM}
\label{sec:examm}

\begin{figure*}[t]
    \centering
	\includegraphics[width=\textwidth]{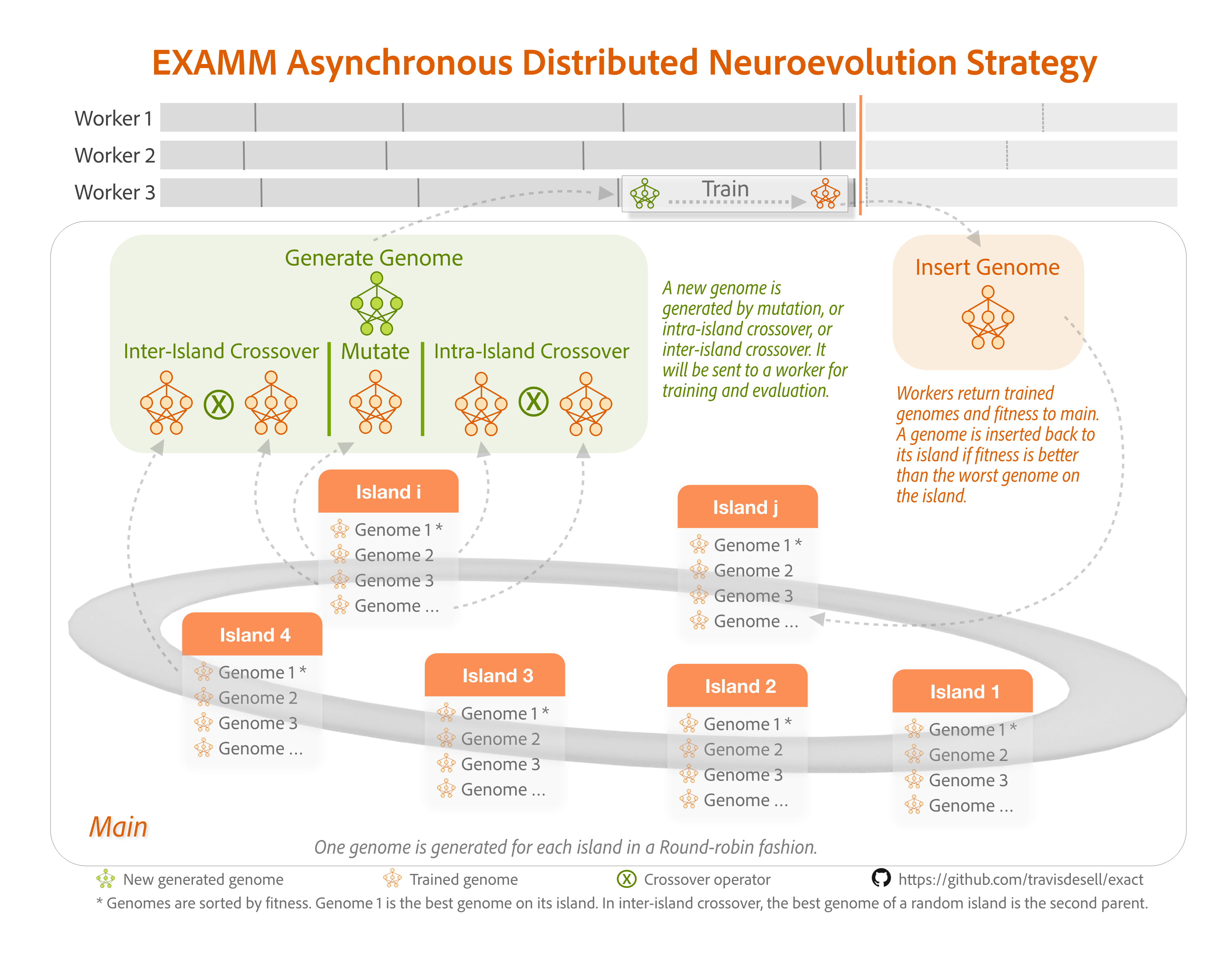}
	\caption{EXAMM flowchart }
	\label{fig:examm}
\end{figure*}

Figure~\ref{fig:examm}~provides a high-level flowchart for the EXAMM algorithm. EXAMM is an offline, distributed asynchronous neuro-evolutionary (NE) algorithm that evolves progressively larger RNNs for large-scale, multivariate, real-world TSF~\cite{elsaid2020improving,elsaid2020neuro}. EXAMM has $n$ islands and each island has a capacity of $m$. Each island starts with a minimum seed genome that only has input-to-output connections. One genome is generated for each island in a round-robin fashion until the entire population reaches the total number of the generated genomes. On each island, the new genome could be generated by mutation, intra-island crossover, or inter-island crossover, in which islands could exchange genes. The generated genome is sent to the next available worker process for training. After training is completed, the trained genome is sent back to the island where it was generated, potentially replacing the worst genome if its fitness is higher than the (current) worst fitness value.

EXAMM evolves RNN architectures consisting of varying degrees of recurrent connections and memory cells through a series of mutation and crossover (reproduction) operations. Memory cells are selected from a neural library including $\Delta$-RNN units~\cite{ororbia2017diff}, gated recurrent units (GRUs)~\cite{chung2014empirical}, long short-term memory cells (LSTMs)~\cite{hochreiter1997long}, minimal gated units (MGUs)~\cite{zhou2016minimal}, and update-gate RNN cells (UGRNNs)~\cite{collins2016capacity}. ONE-NAS utilizes a similar parallel asynchronous strategy which naturally loads balances itself and allows for the decoupling of population size from the number of workers during each generation~\cite{ororbia2019examm}. Generated offspring inherit their weights from their parents, which can significantly reduce the time needed for their training and evaluation~\cite{lyu2021experimental}. It has been shown that EXAMM can swiftly adapt RNNs in transfer learning scenarios, even when the input and output data streams are changed~\cite{elsaid2020improving}~\cite{elsaid2020neuro}. This serves as a preliminary motivation and justification for being able to adapt and evolve RNNs for online TSF.

EXAMM has also evolved RNNs for time series prediction for different real-world applications~\cite{lyu2021neuroevolution, elsaid2020neuro} and performance improvements through the use of extinction and repopulation events/mechanisms, in real-world evolution~\cite{lyu2021Improving}, have been investigated.

\subsection{The ONE-NAS Algorithm}
\label{sec:one_nas}

Algorithms~\ref{alg:onenas}, and~\ref{alg:data} present pseudo-code for the full ONE-NAS procedure and Figure~\ref{fig:onenas}~presents a high-level overview of the asynchronous, distributed, and online ONE-NAS process. 
ONE-NAS concurrently evolves and trains new RNN candidate models while performing online time series data prediction. \emph{Note that ONE-NAS is fully online and does not require pre-training on any historical data before the online NE process begins.}

Figure~\ref{fig:sample}~presents an example of ONE-NAS online prediction performance on multivariate wind turbine dataset. The plotted value is the expected and predicted output parameter \emph{average active power} with values normalized between [0, 1]. We can see that the output values are non-seasonal.

\input{90-pseudocode.tex}

\paragraph{Similarities between EXAMM and ONENAS:} ONE-NAS is also a NE NAS algorithm, so the online learning process typically starts with a minimal seed genome (a minimal seed genomes only has input to output connections and contains no hidden layers). It is also possible to start with a previously trained model or generated architecture as the seed genome to bootstrap this process. Genomes are evolved with the same mutation and crossover methods in EXAMM. The population is maintained by distributed islands, which allows islands to evolve in their own niche, with islands exchanging information only by periodic inter-island crossover. Islands in both methods use a repopulation strategy to periodically erase and repopulate the islands that become stuck in local optima~\cite{lyu2021Improving}, and this work shows that this repopulation is critical for achieving viable online performance in ONE-NAS.

\paragraph{Differences between EXAMM and ONE-NAS:} EXAMM is an offline algorithm, where the evolutionary process happens offline with each generated genome adapted to the entire offline dataset. In contrast, ONE-NAS collects the online streaming data for online training and evaluation, selecting subsets of historical data to be used for training and validation. In this work, crucially, ONE-NAS operates entirely online without any pretraining\footnote{ONE-NAS can be seeded with a prior model or topology - however all results in this work start with an untrained, minimal seed genome}. 

At the end of the neuro-evolutionary process, EXAMM selects the global best genome as the optimal solution. However, in ONE-NAS, the genome generation, training, evaluation, and online prediction processes are all performed online and concurrently. ONE-NAS uses the selected best genome from its previously trained population to make online predictions at each time $t$ while a new population of genomes is dynamically trained. 

During the online evolutionary process, genomes are generated randomly by crossover and mutation, and inevitably, new generations contain genomes that perform worse than the average population. In the online setting, any algorithm needs to produce on-the-fly predictions with an expected accuracy at real-time, so it cannot afford to have too many poorly performing genomes in the population pool, allowing them to be parents to future offspring. ONE-NAS instead selects an elite population from the generated trained genomes and uses the elite population as online predictor candidates and parents for the next generation.


ONE-NAS's online NE process starts with a minimal seed genome, which serves as an initial genome. In ONE-NAS each island has two sub-populations, generated and elite\footnote{EXAMM, on the other hand, operates in a steady-state manner so it does not have explicit populations or elite populations.}. 

At time $t$, ONE-NAS evolves genomes according to the following steps:
\begin{itemize}[noitemsep,nolistsep]
    \item Use the elite population $E_{t-1}$ to generate a set of $m$ genomes through mutation and crossover, defined as the generated population $O_{t}$.
    \item Train the generated genomes $O_{t}$ using MPI\footnote{The Message Passing Interface~\cite{barker2015message}, the most popular high performance computing message passing library.} workers for a specified number of iterations/epochs with randomly selected historical data, $B_{train}$.
    \item Evaluate all of the current generation's genomes, $E_{t-1}$ and $O_{t}$, using recent validation data, $B_{validation}$, to calculate their fitness values.
    \item Select the next elite population, $E_{t}$, from $E_{t-1}$ and $O_{t}$.
    \item Selecting the global best genome from the elite populations, $E_{t}$, from all islands for online prediction.
    \item Retain the members of the elite population $E_{t}$ for the next generation.
\end{itemize} 
Each generation lasts for a specified period $p$, measured in a number of discrete time steps (in this work, $p = 25$), which allows for the processing of a subsequence of the target time series. The best genome from the previous generation performs online predictions of the new subsequence ($B_{next}$) as it arrives while, concurrently, the new generation of genomes is generated and trained. At the end of a generation, this new subsequence of data is added to ONE-NAS's historical training data (memory). During a generation, the generated genomes $O_t$ are trained on a randomly selected set of $B_{train}$ subsequences of historical data, after which the entire population (including the elite $E_t$) will be validated on the most recent $B_{validation}$ subsequences. Each genome's fitness, calculated as the mean squared error loss over $B_{validation}$, is then used to select the next elite population $E_{t+1}$. The best genome in $E_{t+1}$ is used for online prediction in the next generation. 

Note that, while the genomes $O_t$ are trained using backpropagation on batches sampled from the historical data, the RNNs in $E_t$ do not continue to be trained. Since not all RNNs in $E_t$ will perform better than those in $O_t$, ``obsolete'' RNNs will naturally die off over time while RNNs with strong performance will remain/persist. Also note that one of EXAMM's mutation operations, crucially utilized by ONE-NAS, is a clone operation which allows for a duplicate of an elite parent to be retained and trained in the next generation (further re-using its parent's weights due to EXAMM's Lamarckian weight inheritance strategy).

\subsection{Learning Important Information}
\label{sec:important_info}

Online learning models generally suffer from catastrophic forgetting~\cite{hoi2021online} -- this poses a significant problem since preserving important historical temporal information is crucial for TSF problems. One way to counter forgetting and data drift would be to train all of the offspring/genomes on every single historical data point seen so far jointly with any new incoming data. However, this would quickly become inefficient, preventing an adaptive system from operating in a fast and online fashion. In ONE-NAS, offspring are instead trained with new incoming data and only on randomly selected sub-sets of historical data. Historical information is preserved in the population by retaining elite individuals as well as children inheriting weights from those parents~\cite{lyu2021experimental} in tandem with efficient training using only small amounts of randomly selected/stored historical data. The fitness of the trained offspring is evaluated via validation mean squared error (MSE) using the most recent data. This process results in genomes that contain less important temporal information that naturally die off across generations of evolution.

%% file: 90-pseudocode.tex
\begin{algorithm}[H]
    \begin{algorithmic}[1]
        \footnotesize
        \Function{ONE-NAS MAIN}{}

            \For{$t$ in $generations$}
                \LeftComment{\emph{Perform predictions concurrently in a new thread,}}
                \LeftComment{\emph{which returns the next time subsequence when complete}}
                \State $B_{next}$ = $globalBestGenome$.onlinePredict() 
                
                \If{islandRepopulation}
                    \If{t \% extinctFreq == 0}
                    \State $worstIsland$ = rankIslands().last()
                    \State repopulate($worstIsland$, $globalBestGenome$)
                    \EndIf
                \EndIf 
                
                \LeftComment{\emph{Generate genomes in main process}}
                \State $O_t$ = MPIMain.generateGenomes($E_t$)
                \LeftComment{\emph{Process genomes asynchronously in parallel on worker processes}}
                \State $B_{validation}$ = getValidationData($t$, $numValidationSets$)
                \For{genome $g$ in $O_t$}
                    \LeftComment{\emph{Each worker randomly selects different training data}}
                    \State $B_{train}$ = getTrainingData($t$, $numTrainingSets$)
                    \State MPIWorker.trainGenomes($g$, $B_{validation}$, $B_{train}$)
                \EndFor
                \State MPI-Barrier()
                
                \LeftComment{\emph{Evaluate genomes for next generation}}
                \State $E_{t+1}$ = selectElite($E_t$, $O_t$, $B_{validation}$)
                \State $globalBestGenome$ = getGlobalBestGenome()
                

                \State $wait$($B_{next}$) 
                
                \LeftComment{\emph{Get the latest subsequence of data and add it to the historical data}}
                \State timeSeriesSets.append($B_{next}$)
            \EndFor
        \EndFunction
    \end{algorithmic}
    \caption{\label{alg:onenas} {ONE-NAS}}
\end{algorithm}

\begin{algorithm}[h]
    \begin{algorithmic}[1]
    \footnotesize
        \Function{GetTrainingData}{$t$}
            \State timeSeriesSets[0 : $t-numValidationSets$].shuffle()
            \State $B_{train}$ = timeSeriesSets[0 : $numTrainSets$]
        \EndFunction
        
        \Function{GetValidationData}{$t$, $numValidationSets$}
            \State $B_{validation}$ = timeSeriesSets[$t-numValidationSets$ : $t$]
        \EndFunction
        
        \Function{GetOnlineTestData}{$t$, $numValidationSets$}
            \State $B_{next}$ = timeSeriesSets[$t+1$]
        \EndFunction
    \end{algorithmic}
    \caption{\label{alg:data} {Data Selection Methods}}
\end{algorithm}

\label{sec:one-nas}

%% file: 05-results.tex
\begin{figure*}[t]
    \centering
	\includegraphics[width=\textwidth]{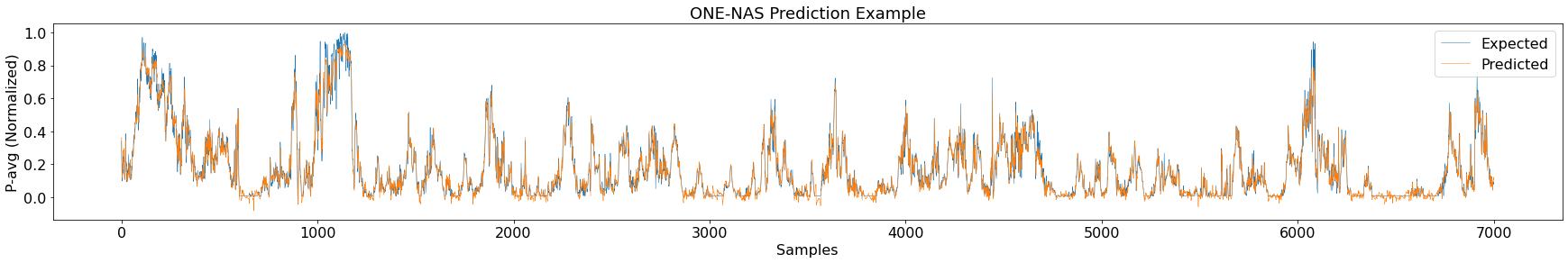}
	\caption{The \emph{average active power} parameter from the wind dataset used in this work, as well as an example of ONE-NAS's predictions on this dataset. As the time series is very long (over $59,000$ datapoints), for visibility, this figure depicts a selection of $7,000$ data points from February to April 2017.}
	\label{fig:sample}
\end{figure*}



\section{Experimental Design}
\label{sec:experiments}

\subsection{Datasets}
\label{sec:datasets}
This work utilized two real-world datasets for predicting time series data with RNNs. The first was wind turbine engine data collected and made available by ENGIE's La Haute Borne open data windfarm\footnote{https://opendata-renewables.engie.com}, which was gathered from 2017 to January 2018. This wind dataset is very long, multivariate (containing $22$ parameters/dimensions), non-seasonal, and the parameter recordings are not independent. The wind turbine data consists of readings every $10$ minutes from 2017 to January 2018. \emph{Average Active Power} was selected as the output parameter to forecast for the wind turbine dataset. Figure~\ref{fig:sample}~provides an example of the noisiness and complexity of the output parameter as well as an example of the accuracy of ONE-NAS's predictions. This entire time series has around $59,000$ data points, with the above example showing $7000$ data points from February to April 2017 as an example. Note that ONE-NAS is trained online (with no pre-training) so this plot depicts online predictive performance as our system learns from scratch. The second dataset is the daily index of the Dow Jones Industrial Average (DJIA) from the years 1885-1962, which is univariate and contains $35,701$ samples. 
Both of the datasets that we investigate contain raw and abnormal data points that have not been cleaned -- spikes and outliers have not been removed or smoothed from either dataset.

\subsection{Processing and Setup}
\label{sec:setup}
Research has shown that using shorter subsequences of time series data during training can improve an RNN's convergence rate as well as its overall performance~\cite{lyu2021neuroevolution}. For these experiments, the original datasets were divided into subsequences of $25$ timesteps each. During each simulated ONE-NAS generation, each newly generated genome was trained on $600$ randomly selected subsequences from the historical data pool and then validated using the most recent $100$ subsequences of data. Each genome utilizes a different random selection of $600$ subsequences. There is no overlap between training and validation data (the most recent $100$ subsequences are added into the historical pool after being used for validation). All the experiments for the wind dataset were run for $2000$ generations, which represents one singular pass over the full wind data time series. 

In the ONE-NAS simulations\footnote{https://github.com/travisdesell/exact}, during each generation, $50$ elite genomes from the previous generation were retained and the elite genomes were used to generate $100$ new genomes using a mutation rate of $0.4$ and crossover probability of $0.6$. Each of the $100$ non-elite genomes in the new generation were trained in a worker process for $10$ epochs of backprop (a local search), with the first $5$ epochs involving training on the original subsequence data. Then, in each of the last $5$ epochs, $10\%$ Gaussian noise was added (using the mean and standard deviation of the sliced data) as an augmentation technique to prevent overfitting~\cite{bishop1995training}~\cite{bishop1995neural}. ONE-NAS with island repopulation utilized $10$, $20$, $30$, or $40$ islands,  each having its own elite population of $5$ genomes which generated an additional $10$ genomes per generation. New genomes were generated with a mutation rate of $0.3$, inter-island crossover rate of $0.4$, and intra-island crossover rate of $0.3$. 


\subsection{Results}
\label{sec:results}

Each experiment was repeated $10$ times using Rochester Institute of Technology's research computing systems. This system consists of $2304$ Intel® Xeon® Gold $6150$ CPU $2.70$GHz cores and $24$ TB RAM, with compute nodes running the RedHat Enterprise Linux 7 system. Each experiment utilized $16$ cores.

\subsubsection{Comparison with Classical TSF Methods}
To test the performance of ONE-NAS, we first compared it to classical TSF methods: naive prediction, moving average prediction, and simple exponential smoothing. While these are very simple univariate methods, they do perform well in many real-world application scenarios, due to the fact that the TSF data is very noisy and using previous observations to estimate the expected value at next time step usually gives a reasonable prediction. Notably, these methods are capable of outperforming complex and sophisticated methods across a wide variety of datasets~\cite{makridakis2018statistical}.


Na\"ve prediction (Naive) simply uses the data's/parameter's previous value, $x_{t-1}$ as the predicted value, $\hat{y_t}$, for the next time step: $\hat{y_t} = x_{t-1}$. The moving average~\cite{cryer1986time} predictor (MA) uses the average of the last $n$ time steps as the prediction of the next time step, where $n$ is the moving average data smoothing window (a hyperparameter): $\hat{y_t} = 1/n * \sum_{1}^{n} x_{t-n}$. The simple exponential smoothing (Holt linear)~\cite{gardner1985exponential} predictor (EXP) computes a running average of the previously seen parameters, where $\alpha$ is the smoothing factor, and $0< \alpha <1$: $\hat{y_t} = \alpha * x_{t-1} + (1 - \alpha) * \hat{y}_{t-1}$.

\begin{figure*}[t]
    \centering
    \includegraphics[width=0.8\textwidth]{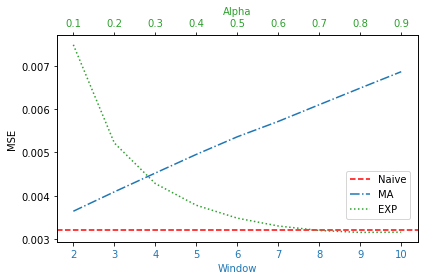}
    \caption{Classical TSF methods with varying window sizes and alpha values.}
    \label{fig:classic-alpha}

\end{figure*}

We are aware that the choice of window size and $\alpha$ significantly affect the prediction performance of the moving average and exponential smoothing methods. Figure~\ref{fig:classic-alpha}~shows the MA and EXP prediction mean squared error (MSE) with different window sizes ($n$) and $\alpha$ values on the wind dataset. The plot shows that the wind dataset is highly complex, where na\"ve almost entirely predicts better than MA or EXP (apart from a negligible improvement with EXP for $alpha$ values of $0.8$ and $0.9$).

\begin{figure*}[t]
    \centering
    \includegraphics[width=0.8\textwidth]{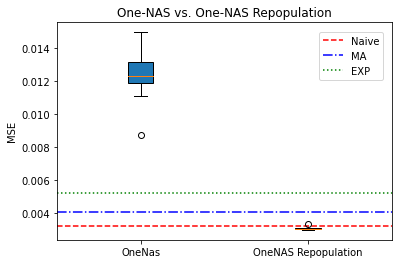}
    \caption{Online prediction MSE of ONE-NAS and three classical TSF methods.}
    \label{fig:onenas-classic}
\end{figure*}

Figure~\ref{fig:onenas-classic}~shows a plot for the online prediction MSE of ONE-NAS. The horizontal lines show each of the three classical time series forecasting methods. Note that these methods are not stochastic, so their performance is always the same. Moving average (MA) predicts with a window size $n = 3$ and exponential smoothing uses a $\alpha = 0.2$. The difference between ``OneNas'' and ``OneNas Repopulation'' on the plots is that ``OneNas'' does not use any islands, where the entire population is one island, and the ``OneNas Repopulation'' uses multiple islands and the island repopulation strategies. The ONE-NAS repopulation strategy shown in this plot used $20$ islands (each with $5$ elite genomes and $10$ other genomes per generation), with an extinction and repopulation frequency of $200$ generations. While the ONE-NAS online prediction without repopulation performs worse compared to all three classical methods, the ONE-NAS method with repopulation not only significantly outperforms the baselines but exhibits better predictive performance than the classical methods across all repeats except for one outlier.

\begin{figure*}[t]
    \centering
    \includegraphics[width=0.8\textwidth]{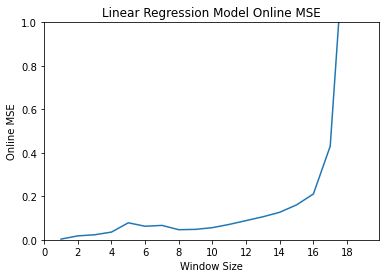}
    \caption{Linear regression window size vs. MSE.}
    \label{fig:linear_regression}
\end{figure*}

We further investigated using linear regression for online learning. At time $t$, the linear regression model is built using the last $n$ observations and is then used to make predictions for time $t+1$. Figure~\ref{fig:linear_regression}~shows the average linear regression online prediction mean squared error (MSE) using different observation window sizes $n$ on the wind dataset. The plot shows that, for this dataset, the linear regressor makes better predictions when the observation window $n$ is smaller. The best results come from the window size of $1$, which is simply the na\"ve prediction from before. The inability of these methods to outperform na\"ve prediction highlights the complexity of this dataset -- simpler prediction methodologies cannot capture the complexity of forecasting this data. Due to this, we focus on using na\"ve prediction MSE to represent classical, baseline prediction performance in the following sections/experiments.

\subsubsection{Preserving Population Variety with Islands}

As shown in Figure~\ref{fig:onenas-classic}, using islands to maintain the population can significantly improve/boost the online forecasting performance, allowing ONE-NAS to outperform the classical methods. It is crucial to preserve variety in an online setting, because: 
\textbf{1)} ONE-NAS generates more genomes than EXAMM (for the same dataset and experimental set up, only 8k genomes were generated using EXAMM, but 200k generated in total using ONE-NAS), 
\textbf{2)} genomes generated online are trained and evaluated with significantly less data (only a subset of the historical data), and, 
\textbf{3)} the elite population is evaluated using the most recent data, so there is some data drift present in each generation, meaning that preserving variety can preserve important historical information while preventing overfitting.


\begin{figure}[t]
    \centering
         \includegraphics[width=0.8\textwidth]{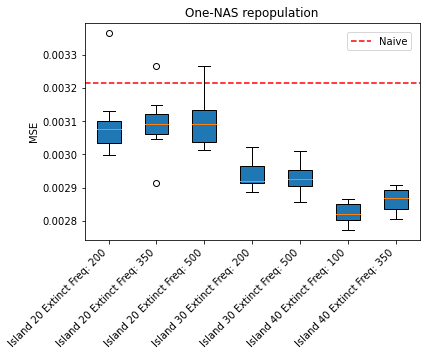}
    	\caption{ONE-NAS MSE with varying island sizes and extinction rates.}
    	\label{fig:onenas-islands}
\end{figure}

\begin{figure}[t]
    \centering
        \includegraphics[width=0.8\textwidth]{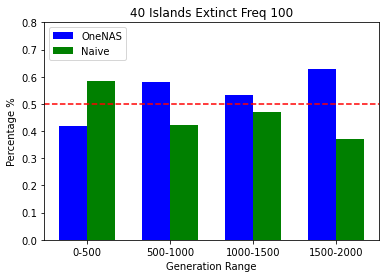}
    	\caption{ONE-NAS percentage of better prediction generations versus classical methods.}
    	\label{fig:onenas-percentage}
\end{figure}

Given the above result that it is possible to effectively evolve and train RNNs in an online setting, we found that two hyperparameters significantly affect online prediction performance in ONE-NAS: the island size and the extinction and repopulation frequency. Figure~\ref{fig:onenas-islands}~shows a box plot of the online prediction MSE using island sizes of $20$, $30$, and $40$ in $10$ repeated experiments/trials with varying repopulation frequencies. As the number of islands increases, the prediction performance improves. This could be due to the fact that more islands allow for more speciation and a greater ability to escape local optima. Additional species also provide more robustness to noise and overfitting of the data. Given the same number of islands, more frequent extinction and repopulation events yield better performance on average. This is most likely due to the fact that these events prevent islands from stagnating with poor(er) species.

\subsubsection{Online Predictions over Time}
The previous plots show the overall performance of ONE-NAS against classical methods for the entire wind time series, which gives an advantage to classical methods since they do not require any training and thus have a significant advantage for earlier time steps when ONE-NAS has not had much opportunity to train/evolve RNNs. In order to investigate how much the RNNs evolved by ONE-NAS were improving as they processed more data in an online fashion, we measured how many time steps (for each generation) exhibited a better predictive performance between the na\"ve method and ONE-NAS. Figure~\ref{fig:onenas-percentage}~shows the percentage of predictions of each method that were accurate as the simulation progressed, with the red line representing $50$\%. This plot shows that, while ONE-NAS does not outperform the na\"ve strategy within the first $500$ generations, the performance of ONE-NAS continues to increase as the evolution continues, which means that ONE-NAS does not only train and predict values online, it also gets progressively better throughout the evolutionary process, which is what we would expect from an online algorithm. Also note that it would be beneficial to combine a classical method with ONE-NAS -- it could prove fruitful to use a na\"ive predictor until ONE-NAS has had enough evolution time to produce more accurate predictions.


\subsubsection{Online Prediction Time Efficiency}
Another key concern for the evaluation of online algorithms, apart from predictive accuracy, is time efficiency.  If an online algorithm cannot provide predictions at a rate less than the arrival rate of new data to be predicted, then it is not usable/viable. ONE-NAS resolves this issue by utilizing the previous best genome to provide predictions while concurrently training the next generation.

\begin{table}
    \centering
    \begin{tabular}{|c|c|c|} \hline

Num & Avg & Longest \\
Islands & Time (s) & Time (s) \\ \hline \hline
        20 & 35.67 & 109.20 \\ \hline
        30 & 41.72 & 127.44  \\ \hline
        40 & 69.96 & 189.72  \\ \hline

    \end{tabular}
    \caption{\label{table:time} The average and longest measured times, in seconds (s), needed to evolve each generation for the wind dataset. 
}
\vspace{-0.55cm}
\end{table}

Note that for this study, each generation was generated and trained over a single subsequence of $25$ time steps.  For the wind dataset, each time step was gathered at a $10$ minute interval, so this provides a significant buffer. However, for many time series datasets, time step frequency can be per minute, per second, or even faster - making time efficiency a serious concern. Table~\ref{table:time} presents the average and longest time required to evolve and train one generation of the ONE-NAS Repopulation experiments for the varying numbers of islands. Note that population size was tied to island size, with $5$ elite genomes and $10$ other genomes per island, which is why the $40$ island genomes took approximately twice the time. In the worst case, for $40$ islands, the longest time per generation was a bit above $3$ minutes, which is far below the $250$ minute generation time for the wind data.  

It should also be noted that the workers training the genomes for each generation were distributed across $16$ processors and that performance will scale linearly upwards until the number of available processors is equal to the population size (i.e., all generated genomes can be independently trained in parallel without reduction in performance apart from a fixed communication and genome generation overhead cost). This particular scalability of ONE-NAS makes it well-suited to online learning. For example, if we scaled up to $200$ processors over the $16$ used for the $20$ island experiments, we can estimate approximately $2.85$ seconds per generation (\ie, training the $200$ non-elite genomes at once, instead of $16$ at a time), plus some additional communication and generation overhead. With a generation time of $25$ time steps, this would allow for incoming data to be processed at almost $10$ readings per second. Given some flexibility in determining subsequence/generation time, ONE-NAS demonstrates the potential to operate for very high frequency time series given enough computing power.

\begin{figure}[t]
    \centering
    \includegraphics[width=0.8\textwidth]{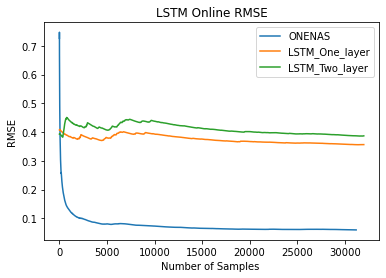}
    \caption{ONE-NAS versus one and two layer online LSTM RNNs.}
    \label{fig:lstm}
\end{figure}
\begin{figure}[t]
    \centering
    \includegraphics[width=0.8\textwidth]{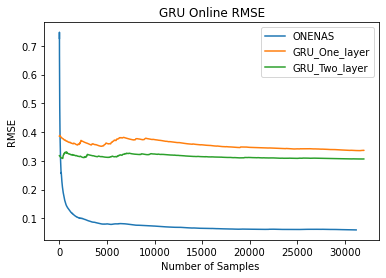}
    \caption{ONE-NAS versus one and two layer online GRU RNNs.}
    \label{fig:gru}
\end{figure}

\subsubsection{ONE-NAS vs LSTM \& GRU RNNs}
\label{sec:onenas_vs_lstm_grus}
LSTM and GRU RNNs have been widely used in TSF problems~\cite{chimmula2020time}~\cite{fu2016using}. 
Fixed one-layer and two-layer LSTM and GRU networks were trained online to compare with ONE-NAS, where one-layer indicates that an LSTM/GRU only contained one fixed hidden layer or neurons while two-layer indicates that the RNN contained two fixed hidden layers. The size of each hidden layer was set to be equal to the size of the input layer and each node/unit in the hidden layer was set to be an LSTM or GRU cell (whereas the input and output layer nodes consisted of simple neurons). Note that each layer is fully connected to the next. The LSTM and GRU networks were initialized with uniform random $\mathcal{U}(-0.5, 0.5)$ weights and weight updates/gradients were applied using Nesterov's accelerated gradient with a momentum value of $\mu = 0.9$. 

The LSTM and GRU networks were trained on the wind data with a fixed moving window size. The gradients for the fixed-layer RNNS were all computed online using truncated backpropagation through time~\cite{marschall2020unified}~\cite{gers2002applying} (where the networks were unrolled over the length of the window in order to compute the full gradients). 
Figure~\ref{fig:lstm}~and~\ref{fig:gru} show online testing RMSE using the best-found window size and best-found learning rate over $10$ runs. A window size of $30$ and a learning rate of $1e^{-3}$ was used for both one-layer LSTM and GRU networks, a window size of $20$ and learning rate of $1e^{-3}$ was used for for the two-layer LSTM network, and a window size of $20$ and learning rate of $2e^{-3}$ was used for the two-layer GRU network. These configurations were selected from experiments with different fixed window size of $n = 10, 20, 30, 40, 50$, and a variety of potential learning rates ranging from $5e^{-3}$ to $1e^{-4}$. The results show that ONE-NAS significantly outperforms tuned, fixed one-layer and two-layer LSTM and GRU networks.




\begin{figure}[t]
    \centering
    \includegraphics[width=0.8\textwidth]{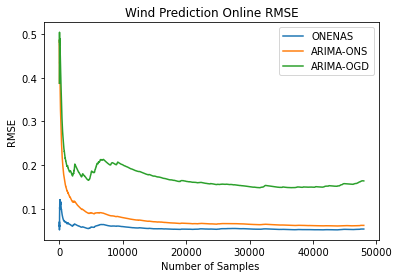}
    \caption{Online ARIMA versus ONE-NAS predictions on the wind dataset.}
    \label{fig:onenas-arima-wind}
\end{figure}
\begin{figure}[t]
    \centering
    \includegraphics[width=0.8\textwidth]{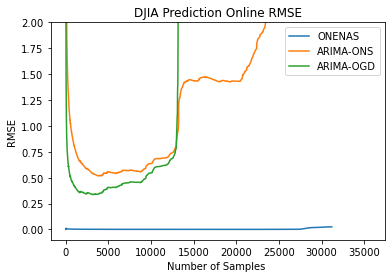}
    \caption{Online ARIMA versus ONE-NAS predictions on the DIJA dataset.}
    \label{fig:onenas-arima-djia}
\end{figure}

\subsubsection{ONE-NAS vs Online ARIMA}

While there is a significant lack of methods for online multivariate TSF, recent work by Liu \etal~\cite{liu2016online} has led to the development of an online ARIMA method for univariate TSF. To compare ONE-NAS with a state-of-the-art method as opposed to only the classical methods investigated earlier, we investigate ONE-NAS performance alongside this powerful, online ARIMA model. To reproduce the results from Liu \etal, we first performed experiments using the Dow Jones Industrial Dataset (DJIA), which was used for evaluation in their work. Figure~\ref{fig:onenas-arima-djia} presents results for their ARIMA-ONS (Arima Online Newton Step) and ARIMA-OGD (ARIMA Online Gradient descent) variants, which were the best performing variants examined in the original study. These were compared to the ONE-NAS Repopulation method with $10$ islands and extinction frequency of $200$, and over a similar generation and subsequence length of $25$. For both ONE-NAS and online ARIMA, the plots show that the online root mean squared error (RMSE) over time, averaged over $10$ repeated experiments. The online RMSE over time is calculated as the average RSME of all previous predictions. For this DJIA data, we show that, although the online ARIMA predictor mirrors/reproduces the results from the original study, ONE-NAS clearly outperforms this method by multiple orders of magnitude. 

We finally compared ONE-NAS with online ARIMA on the wind datasets, the results of which are presented in Figure~\ref{fig:onenas-arima-wind}. Similarly, the results depict the average performance over $10$ repeated experiments.  For online ARIMA, we performed a hyper-parameter sweep using a grid search for the online ARIMA methods on its learning rates and $\epsilon$ and report the best found hyper-parameters. For ARIMA-ONS, the learning rate was set to $e^{-3}$ and $\epsilon = 3.16e^{-6}$. For ARIMA-OGD, learning rate was set to $e^3$, and $\epsilon = e^{-5.5}$. ONE-NAS used the best hyper-parameters from our earlier previous results/experiments in this study ($40$ islands and extinction frequency of $100$). Similarly, we find that ONE-NAS also significantly outperforms the online ARIMA methods on the wind dataset.

\subsection{ONE-NAS Evolved RNNs}
\label{sec:evolved_rnn_example}

\begin{sidewaysfigure}
\includegraphics[width=0.95\textwidth,height=0.6\textheight]{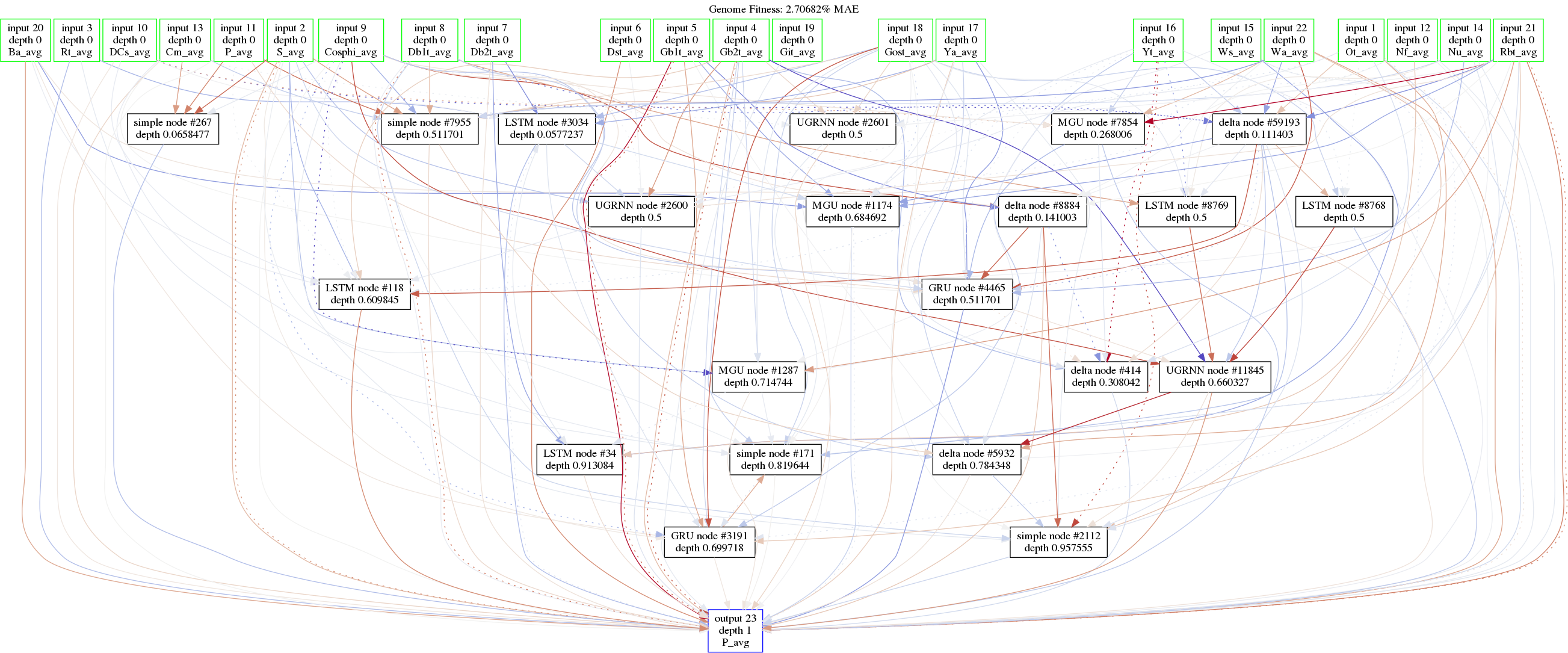}
\caption{An examplar best performing RNN that was evolved online by ONE-NAS.}
\label{fig:evolved_rnn}
\end{sidewaysfigure}

Figure~\ref{fig:evolved_rnn} presents an example RNN evolved by ONE-NAS. In comparison to commonly used layer-based (hierarchical) neural networks, networks evolved by ONE-NAS are ``unlayered'' but exhibit highly complex connectivity structures.  Nodes in the network show their selected memory cell type (or simple, if a basic neuron was chosen), and edges with positive weights are shown in blue, and edges with negative weights are shown in red.  Feed forward connections are in solid lines, and recurrent connections are shown in dotted lines.

While this network may seem complex, in comparison to a standard layer based GRU or LSTM network (which ONE-NAS was shown to outperform in Section~\ref{sec:onenas_vs_lstm_grus}), the evolved network only has $21$ neurons (of varying types). Whereas the GRU and LSTM networks have hidden layers of size $22$ (the same size as the input layer). Furthermore, the nodes in the evolved network are not fully connected and, as a result, actually contains significantly less synaptic edges. This example shows how ONE-NAS not evolves well performing architectures online but also ones that are sparser (with respect to synaptic connectivity) and computationally more efficient.

%% file: 06-conclusion.tex
\section{Conclusion}
\label{sec:conclusion}

This work presents the Online NeuroEvolution-based Neural Architecture Search (ONE-NAS) meta-heuristic optimization algorithm and applies it to the problem of non-stationary time series forecasting (TSF) on challenging real-world tasks. To the author's knowledge, ONE-NAS is the first neural architecture search algorithm capable of designing and training recurrent neural networks (RNNs) in real-time as data arrives in an online fashion. ONE-NAS is a dynamic/online, distributed, scalable, real-time algorithm that works on univariate and multivariate real-world TSF datasets. ONE-NAS starts evolution from a minimal seed genome, which potentially reduces optimal model complexity~\cite{stanley2002evolving}, and then generates, trains, evaluates genomes online/incrementally, while concurrently performing online forecasting with the best previously found model. New streaming data is collected into a historical data pool and new generated genomes are trained on randomly selected sub-sets of the historical data. Generated genomes retain knowledge from parental weights using a Lamarckian inheritance process~\cite{lyu2021experimental}, reducing the amount of training required. By training new genomes with randomly selected historical data and evaluating these on recently collected data, data drift can be managed and catastrophic forgetting can be avoided. Maintaining genomes in populations also acts as a method to retain previously gained knowledge on historical data to further safeguard against forgetting.

An important feature of the ONE-NAS algorithm is that it utilizes islands to maintain the population diversity and  prevent over-fitting during the online learning process. Further, periodically repopulating poorly performing islands was shown to be critical in allowing ONE-NAS to outperform other strategies. Our empirical results show that ONE-NAS repopulation outperforms classical TSF methods, linear regression models, LSTM and GRU networks trained online, and a powerful online ARIMA method. Our results also demonstrate that using more islands with more frequent repopulation is strongly correlated with increasing/improved performance. Finally, our statistical results indicate that our algorithm can achieve real-time performance in real-world scenarios.  As a result, this study shows that online neuroevolution or neural architecture search is feasible in online scenarios, which holds great promise for addressing important challenges in time series forecasting.

%% file: 00-main.bbl
\begin{thebibliography}{10}
\expandafter\ifx\csname url\endcsname\relax
  \def\url#1{\texttt{#1}}\fi
\expandafter\ifx\csname urlprefix\endcsname\relax\def\urlprefix{URL }\fi
\expandafter\ifx\csname href\endcsname\relax
  \def\href#1#2{#2} \def\path#1{#1}\fi

\bibitem{zinouri2018modelling}
N.~Zinouri, K.~M. Taaffe, D.~M. Neyens, Modelling and forecasting daily
  surgical case volume using time series analysis, Health Systems 7~(2) (2018)
  111--119.

\bibitem{wu2012online}
T.~Wu, K.~Xie, D.~Xinpin, G.~Song, A online boosting approach for traffic flow
  forecasting under abnormal conditions, in: 2012 9th International Conference
  on Fuzzy Systems and Knowledge Discovery, IEEE, 2012, pp. 2555--2559.

\bibitem{cao2019financial}
J.~Cao, Z.~Li, J.~Li, Financial time series forecasting model based on ceemdan
  and lstm, Physica A: Statistical Mechanics and its Applications 519 (2019)
  127--139.

\bibitem{lyu2021neuroevolution}
Z.~Lyu, S.~Patwardhan, D.~Stadem, J.~Langfeld, S.~Benson, S.~Thoelke,
  T.~Desell, Neuroevolution of recurrent neural networks for time series
  forecasting of coal-fired power plant operating parameters, in: Proceedings
  of the Genetic and Evolutionary Computation Conference Companion, 2021, pp.
  1735--1743.

\bibitem{guo2016robust}
T.~Guo, Z.~Xu, X.~Yao, H.~Chen, K.~Aberer, K.~Funaya, Robust online time series
  prediction with recurrent neural networks, in: 2016 IEEE International
  Conference on Data Science and Advanced Analytics (DSAA), Ieee, 2016, pp.
  816--825.

\bibitem{fields2019mitigating}
T.~Fields, G.~Hsieh, J.~Chenou, Mitigating drift in time series data with noise
  augmentation, in: 2019 International Conference on Computational Science and
  Computational Intelligence (CSCI), IEEE, 2019, pp. 227--230.

\bibitem{mccloskey1989catastrophic}
M.~McCloskey, N.~J. Cohen, Catastrophic interference in connectionist networks:
  The sequential learning problem, in: Psychology of learning and motivation,
  Vol.~24, Elsevier, 1989, pp. 109--165.

\bibitem{french1999catastrophic}
R.~M. French, Catastrophic forgetting in connectionist networks, Trends in
  cognitive sciences 3~(4) (1999) 128--135.

\bibitem{raju2020iot}
M.~P. Raju, A.~J. Laxmi, Iot based online load forecasting using machine
  learning algorithms, Procedia Computer Science 171 (2020) 551--560.

\bibitem{partee2022using}
S.~Partee, M.~Ellis, A.~Rigazzi, A.~E. Shao, S.~Bachman, G.~Marques,
  B.~Robbins, Using machine learning at scale in numerical simulations with
  smartsim: An application to ocean climate modeling, Journal of Computational
  Science 62 (2022) 101707.

\bibitem{gonzalez2019fuzzy}
J.~A.~R. Gonz{\'a}lez, J.~F. Sol{\'\i}s, H.~J.~F. Huacuja, J.~J.~G. Barbosa,
  R.~A.~P. Rangel, Fuzzy ga-svr for mexican stock exchange's financial time
  series forecast with online parameter tuning, International Journal of
  Combinatorial Optimization Problems and Informatics 10~(1) (2019) 40.

\bibitem{lyu2022one}
Z.~Lyu, T.~Desell, One-nas: An online neuroevolution based neural architecture
  search for time series forecasting, arXiv preprint arXiv:2202.13471.

\bibitem{hoi2021online}
S.~C. Hoi, D.~Sahoo, J.~Lu, P.~Zhao, Online learning: A comprehensive survey,
  Neurocomputing 459 (2021) 249--289.

\bibitem{yu2007online}
L.~Yu, S.~Wang, K.~K. Lai, An online learning algorithm with adaptive
  forgetting factors for feedforward neural networks in financial time series
  forecasting, Nonlinear dynamics and systems theory 7~(1) (2007) 51--66.

\bibitem{cryer1986time}
J.~D. Cryer, Time series analysis, Vol. 286, Springer, 1986.

\bibitem{liu2016online}
C.~Liu, S.~C. Hoi, P.~Zhao, J.~Sun, Online arima algorithms for time series
  prediction, in: Thirtieth AAAI conference on artificial intelligence, 2016.

\bibitem{lyu2021experimental}
Z.~Lyu, A.~ElSaid, J.~Karns, M.~Mkaouer, T.~Desell, An experimental study of
  weight initialization and lamarckian inheritance on neuroevolution, The 24th
  International Conference on the Applications of Evolutionary Computation
  (EvoStar: EvoApps).

\bibitem{lyu2021Improving}
Z.~Lyu, J.~Karnas, A.~ElSaid, M.~Mkaouer, T.~Desell, Improving distributed
  neuroevolution using island extinction and repopulation, The 24th
  International Conference on the Applications of Evolutionary Computation
  (EvoStar: EvoApps).

\bibitem{stine1987estimating}
R.~A. Stine, Estimating properties of autoregressive forecasts, Journal of the
  American statistical association 82~(400) (1987) 1072--1078.

\bibitem{gardner1985exponential}
E.~S. Gardner~Jr, Exponential smoothing: The state of the art, Journal of
  forecasting 4~(1) (1985) 1--28.

\bibitem{hansen1970triple}
J.~I. Hansen, Triple exponential smoothing; a tool for common stock price
  prediction.

\bibitem{siregar2017comparison}
B.~Siregar, I.~Butar-Butar, R.~Rahmat, U.~Andayani, F.~Fahmi, Comparison of
  exponential smoothing methods in forecasting palm oil real production, in:
  Journal of Physics: Conference Series, Vol. 801, IOP Publishing, 2017, p.
  012004.

\bibitem{chen2019state}
Z.~Chen, Q.~Xue, R.~Xiao, Y.~Liu, J.~Shen, State of health estimation for
  lithium-ion batteries based on fusion of autoregressive moving average model
  and elman neural network, IEEE access 7 (2019) 102662--102678.

\bibitem{benvenuto2020application}
D.~Benvenuto, M.~Giovanetti, L.~Vassallo, S.~Angeletti, M.~Ciccozzi,
  Application of the arima model on the covid-2019 epidemic dataset, Data in
  brief 29 (2020) 105340.

\bibitem{kozitsin2021online}
V.~Kozitsin, I.~Katser, D.~Lakontsev, Online forecasting and anomaly detection
  based on the arima model, Applied Sciences 11~(7) (2021) 3194.

\bibitem{schmidt2018unsupervised}
F.~Schmidt, F.~Suri-Payer, A.~Gulenko, M.~Wallschl{\"a}ger, A.~Acker, O.~Kao,
  Unsupervised anomaly event detection for cloud monitoring using online arima,
  in: 2018 IEEE/ACM International Conference on Utility and Cloud Computing
  Companion (UCC Companion), IEEE, 2018, pp. 71--76.

\bibitem{anava2013online}
O.~Anava, E.~Hazan, S.~Mannor, O.~Shamir, Online learning for time series
  prediction, in: Conference on learning theory, PMLR, 2013, pp. 172--184.

\bibitem{han2018multivariate}
M.~Han, S.~Zhang, M.~Xu, T.~Qiu, N.~Wang, Multivariate chaotic time series
  online prediction based on improved kernel recursive least squares algorithm,
  IEEE transactions on cybernetics 49~(4) (2018) 1160--1172.

\bibitem{anava2015online}
O.~Anava, E.~Hazan, A.~Zeevi, Online time series prediction with missing data,
  in: International Conference on Machine Learning, PMLR, 2015, pp. 2191--2199.

\bibitem{yang2017robust}
H.~Yang, Z.~Pan, Q.~Tao, Robust and adaptive online time series prediction with
  long short-term memory, Computational intelligence and neuroscience 2017.

\bibitem{wang2014online}
X.~Wang, M.~Han, Online sequential extreme learning machine with kernels for
  nonstationary time series prediction, Neurocomputing 145 (2014) 90--97.

\bibitem{celik2022online}
B.~Celik, P.~Singh, J.~Vanschoren, Online automl: An adaptive automl framework
  for online learning, arXiv preprint arXiv:2201.09750.

\bibitem{celik2021adaptation}
B.~Celik, J.~Vanschoren, Adaptation strategies for automated machine learning
  on evolving data, IEEE Transactions on Pattern Analysis and Machine
  Intelligence 43~(9) (2021) 3067--3078.

\bibitem{madrid2019towards}
J.~G. Madrid, H.~J. Escalante, E.~F. Morales, W.-W. Tu, Y.~Yu, L.~Sun-Hosoya,
  I.~Guyon, M.~Sebag, Towards automl in the presence of drift: first results,
  arXiv preprint arXiv:1907.10772.

\bibitem{yan2022privacy}
C.~Yan, Y.~Zhang, Q.~Zhang, Y.~Yang, X.~Jiang, Y.~Yang, B.~Wang,
  Privacy-preserving online automl for domain-specific face detection, arXiv
  preprint arXiv:2203.08399.

\bibitem{ororbia2018using}
I.~Ororbia, G.~Alexander, F.~Linder, J.~Snoke, Using neural generative models
  to release synthetic twitter corpora with reduced stylometric identifiability
  of users, arXiv preprint arXiv:1606.01151.

\bibitem{cardamone2010learning}
L.~Cardamone, D.~Loiacono, P.~L. Lanzi, Learning to drive in the open racing
  car simulator using online neuroevolution, IEEE Transactions on Computational
  Intelligence and AI in Games 2~(3) (2010) 176--190.

\bibitem{agogino2000online}
A.~Agogino, K.~Stanley, R.~Miikkulainen, Online interactive neuro-evolution,
  Neural Processing Letters 11~(1) (2000) 29--38.

\bibitem{stanley2005real}
K.~O. Stanley, B.~D. Bryant, R.~Miikkulainen, Real-time neuroevolution in the
  nero video game, IEEE transactions on evolutionary computation 9~(6) (2005)
  653--668.

\bibitem{galassi2016evolutionary}
M.~Galassi, N.~Capodieci, G.~Cabri, L.~Leonardi, Evolutionary strategies for
  novelty-based online neuroevolution in swarm robotics, in: 2016 IEEE
  International Conference on Systems, Man, and Cybernetics (SMC), IEEE, 2016,
  pp. 002026--002032.

\bibitem{stanley2002evolving}
K.~Stanley, R.~Miikkulainen, Evolving neural networks through augmenting
  topologies, Evolutionary computation 10~(2) (2002) 99--127.

\bibitem{sutton2018reinforcement}
R.~S. Sutton, A.~G. Barto, Reinforcement learning: An introduction, MIT press,
  2018.

\bibitem{ororbia2019examm}
A.~Ororbia, A.~ElSaid, T.~Desell,
  \href{http://doi.acm.org/10.1145/3321707.3321795}{Investigating recurrent
  neural network memory structures using neuro-evolution}, in: Proceedings of
  the Genetic and Evolutionary Computation Conference, GECCO '19, ACM, New
  York, NY, USA, 2019, pp. 446--455.
\newblock \href {http://dx.doi.org/10.1145/3321707.3321795}
  {\path{doi:10.1145/3321707.3321795}}.
\newline\urlprefix\url{http://doi.acm.org/10.1145/3321707.3321795}

\bibitem{elsaid2020improving}
A.~ElSaid, J.~Karns, Z.~Lyu, D.~Krutz, A.~Ororbia, T.~Desell, Improving
  neuroevolutionary transfer learning of deep recurrent neural networks through
  network-aware adaptation, in: Proceedings of the 2020 Genetic and
  Evolutionary Computation Conference, 2020, pp. 315--323.

\bibitem{elsaid2020neuro}
A.~ElSaid, J.~Karnas, Z.~Lyu, D.~Krutz, A.~G. Ororbia, T.~Desell,
  Neuro-evolutionary transfer learning through structural adaptation, in:
  International Conference on the Applications of Evolutionary Computation
  (Part of EvoStar), Springer, 2020, pp. 610--625.

\bibitem{ororbia2017diff}
A.~G. Ororbia~II, T.~Mikolov, D.~Reitter,
  \href{https://doi.org/10.1162/neco\_a\_01017}{Learning simpler language
  models with the differential state framework}, Neural Computation 0~(0)
  (2017) 1--26, pMID: 28957029.
\newblock \href {http://arxiv.org/abs/https://doi.org/10.1162/neco\_a\_01017}
  {\path{arXiv:https://doi.org/10.1162/neco\_a\_01017}}, \href
  {http://dx.doi.org/10.1162/neco\_a\_01017}
  {\path{doi:10.1162/neco\_a\_01017}}.
\newline\urlprefix\url{https://doi.org/10.1162/neco\_a\_01017}

\bibitem{chung2014empirical}
J.~Chung, C.~Gulcehre, K.~Cho, Y.~Bengio, Empirical evaluation of gated
  recurrent neural networks on sequence modeling, arXiv preprint
  arXiv:1412.3555.

\bibitem{hochreiter1997long}
S.~Hochreiter, J.~Schmidhuber, Long short-term memory, Neural Computation 9~(8)
  (1997) 1735--1780.

\bibitem{zhou2016minimal}
G.-B. Zhou, J.~Wu, C.-L. Zhang, Z.-H. Zhou, Minimal gated unit for recurrent
  neural networks, International Journal of Automation and Computing 13~(3)
  (2016) 226--234.

\bibitem{collins2016capacity}
J.~Collins, J.~Sohl-Dickstein, D.~Sussillo, Capacity and trainability in
  recurrent neural networks, arXiv preprint arXiv:1611.09913.

\bibitem{barker2015message}
B.~Barker, Message passing interface (mpi), in: Workshop: high performance
  computing on stampede, Vol. 262, Cornell University Publisher Houston, TX,
  USA, 2015.

\bibitem{bishop1995training}
C.~M. Bishop, Training with noise is equivalent to tikhonov regularization,
  Neural computation 7~(1) (1995) 108--116.

\bibitem{bishop1995neural}
C.~M. Bishop, et~al., Neural networks for pattern recognition, Oxford
  university press, 1995.

\bibitem{makridakis2018statistical}
S.~Makridakis, E.~Spiliotis, V.~Assimakopoulos, Statistical and machine
  learning forecasting methods: Concerns and ways forward, PloS one 13~(3)
  (2018) e0194889.

\bibitem{chimmula2020time}
V.~K.~R. Chimmula, L.~Zhang, Time series forecasting of covid-19 transmission
  in canada using lstm networks, Chaos, Solitons \& Fractals 135 (2020) 109864.

\bibitem{fu2016using}
R.~Fu, Z.~Zhang, L.~Li, Using lstm and gru neural network methods for traffic
  flow prediction, in: 2016 31st Youth Academic Annual Conference of Chinese
  Association of Automation (YAC), IEEE, 2016, pp. 324--328.

\bibitem{marschall2020unified}
O.~Marschall, K.~Cho, C.~Savin, A unified framework of online learning
  algorithms for training recurrent neural networks, Journal of machine
  learning research.

\bibitem{gers2002applying}
F.~A. Gers, D.~Eck, J.~Schmidhuber, Applying lstm to time series predictable
  through time-window approaches, in: Neural Nets WIRN Vietri-01, Springer,
  2002, pp. 193--200.

\end{thebibliography}
